\documentclass[twoside]{biorxiv}
\allowdisplaybreaks[1]

\leadauthor{Lauer}

\begin{document}

\title{Learning golf swing signatures from a single wrist-worn inertial sensor}

\author[1,\Letter]{Jessy Lauer}


\maketitle

\begin{abstract}Despite its recognized importance for performance and injury prevention, golf swing analysis remains limited by a reliance on isolated metrics, underrepresentation of professional athletes, and a lack of rich, interpretable movement representations. We address these gaps with a holistic, data-driven framework for personalized golf swing analysis from a single wrist-worn sensor. We build a large dataset of professional swings from publicly available videos, reconstruct full-body 3D kinematics using biologically accurate human mesh recovery, and generate synthetic inertial data to train neural networks that infer motion and segment swing phases from wrist-based input. We learn a compositional, discrete vocabulary of motion primitives that facilitates the detection and visualization of technical flaws, and is expressive enough to predict player identity, club type, sex, and age. Our system accurately estimates full-body kinematics and swing events from wrist data, delivering lab-grade motion analysis on-course and supporting early detection of anomalous movement patterns. Explainability methods reveal subtle, individualized movement signatures, reinforcing the view that variability is a hallmark of skilled performance. Longitudinal tracking demonstrates practical value: as one player's handicap improved from 50 to 2.2 over 1.5 years, our system retrospectively captured measurable technical progress and provided targeted, actionable feedback. Our findings challenge common assumptions—such as swing consistency across clubs and the existence of a single “ideal” swing—and uncover latent biomarkers shaped by both intrinsic traits (e.g., sex, age) and task-specific constraints (e.g., club type). This work bridges lab and field-based biomechanics, offering scalable, accessible, high-fidelity motion analysis for research, coaching, and injury prevention, while opening new directions in movement-based phenotyping, personalized equipment design, and motor skill development.
\end{abstract}



\begin{corrauthor}
    jlauer\at rowland.harvard.edu
\end{corrauthor}

\section*{Introduction}

With nearly 38,000 courses worldwide and an estimated 109.5 million players participating on- and off-course as of 2024~\cite{golfstatsus2024,golfstatsra2024}, golf is one of the most widely played sports globally. Among amateur golfers, performance improvement is a major motivator: in a survey of 600 players, two-thirds cited the desire to improve as a key reason for playing~\cite{theriault1996golf}. Technical refinement of the golf swing—often achieved through visualization and emulation of expert players—is not only critical to performance~\cite{hume2005role,keogh2012evidence}, but also essential for injury prevention~\cite{theriault1998golf}.

Despite increasing interest in golf biomechanics, current research suffers from small sample sizes, underrepresentation of professional and female athletes, methodological inconsistencies, and a narrow focus on a few discrete parameters (e.g., X-factor, crunch factor, clubhead trajectory, etc.;~\cite{bourgain2022golf}). These constraints hinder a more holistic and unbiased understanding of swing mechanics. As high-level coaches have noted, there is a pressing need for advanced biomechanical tools that analyze the swing as a whole~\cite{smith2015golf}, potentially using unsupervised, data-driven methods to uncover underlying motion primitives~\cite{gloersen2018technique}.

Advances in wearable sensing and machine learning, combined with the ubiquity of smartphones and inertial sensors, offer new avenues for scalable, accessible, and high-quality motion analysis~\cite{suo2024motion}. Unlike traditional laboratory setups—which rely on expensive optical motion capture and expert labor—lightweight, portable methods enable cost-effective, naturalistic biomechanical analysis outside the laboratory.

Previous work in this direction has begun to leverage markerless pose estimation and neural representations of swing dynamics. These studies range from basic 2D trajectory tracking of the golfer and club~\cite{yamamoto2023extracting,jiang2022golfpose} to more sophisticated techniques aimed at identifying meaningful patterns in kinematic data. Recent approaches have explored self-supervised learning to detect misalignments between swing sequences~\cite{liao2022ai}, 2D body geometry embeddings to highlight pose discrepancies~\cite{ju2023golfmate}, and lifted 2D poses to 3D to capture richer spatial information~\cite{lee2025golfpose}. Despite these advances, many existing methods fall short in key areas: they often lack physical grounding, making it difficult to interpret the results biomechanically; they disregard temporal coherence, limiting their ability to model the continuous nature of the swing; and they typically require fixed viewpoints or constrained settings, reducing their applicability in real-world scenarios.

In this work, we develop a scalable system for accurate, biomechanically informed golf swing analysis that combines markerless 3D motion capture and discrete motion representation learning. We construct a large-scale database of professional golf swings in natural conditions from publicly available online videos, using state-of-the-art human mesh recovery and high-fidelity musculoskeletal modeling. From these reconstructions, we generate synthetic inertial data from a virtual wrist-worn sensor to train models that predict full-body motion and key swing events from device input. We learn a discrete vocabulary of swing primitives through vector quantization, supporting compact modeling and motion synthesis, and demonstrate its utility in anomaly detection, interpretable feedback, and downstream tasks such as the prediction of player identity, club type, sex, and age. Finally, we illustrate how this representation supports longitudinal tracking of swing improvements over time.

\section*{Methods}

\subsection*{Overview}

We present a hybrid data-driven and physics-informed pipeline to reconstruct full-body 3D golf swing kinematics from a single wrist-worn inertial sensor. The system is trained on synthetic inertial data generated from biomechanically accurate reconstructions of professional swings and leverages motion tokenization to capture compact, discrete representations of individualized swing signatures. This enables not only accurate motion reconstruction, but also downstream analysis of player-specific characteristics. We describe each stage of the pipeline below.

\subsection*{Data curation}

\begin{figure*}[!htbp]
    \centering
    \includegraphics[width=0.9\textwidth]{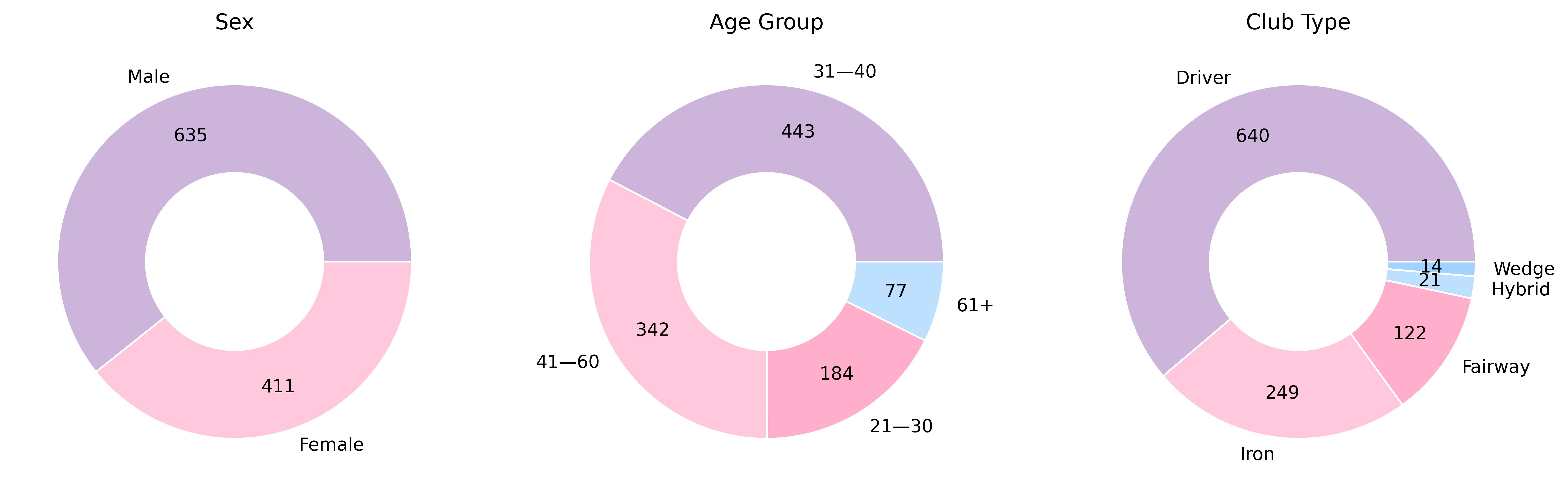}
    \caption{{\bf Distribution of players' sex, age group, and golf club types (\textit{N}=1046).} The dataset comprises a higher proportion of male golfers and driver swings, and a predominance of players aged 31–40.}
    \label{fig:donuts}
\end{figure*}

We extend the GolfDB database—a collection of 758 high-quality, real-time golf swing videos (720p, 30 FPS) of 246 professional golfers from the PGA, LPGA and Champions Tours~\cite{McNally_2019_CVPR_Workshops}—with 288 other YouTube video clips, which include 55 new individuals. Consistent with the original dataset, eight golf events per sequence were carefully annotated (address, toe-up, mid-backswing, top, mid-downswing, impact, mid-follow-through, and finish). Additionally, players' names, sex and age, and golf club types were determined from the video titles and publicly available information. Our dataset therefore comprises a total of 1,046 golf swings (139,503 frames, or approximately 1.3 hours) captured from various viewpoints (e.g., down-the-line, face-on, and rear), with a median number of swings per golfer of 2 (interquartile range [IQR]: 1–4) (Fig.~\ref{fig:donuts}).

\subsection*{Video analysis}

Three-dimensional golf swing kinematics were estimated from monocular videos using WHAM, a fast video-based mesh recovery method that demonstrates state-of-the-art performance on multiple in-the-wild datasets for 3D human pose estimation~\cite{Shin_2024_CVPR}. WHAM is pre-trained on a large-scale dataset of more than 40 hours of motion capture data and 11,000 motion sequences~\cite{Mahmood_2019_ICCV}, and fuses visual information over time, along with motion features, camera movement, and realistic foot contact estimates to reconstruct temporally smooth 3D whole body kinematics with unprecedented accuracy. Specifically, given a raw input video $\{ I^{(t)} \}_{t=0}^T$, WHAM outputs a sequence of SMPL model parameters~\cite{loper2015smpl}, which encode both body pose $\{ \theta^{(t)} \}_{t=0}^T \in \mathbb{R}^{72}$ and shape $\{ \beta^{(t)} \}_{t=0}^T \in \mathbb{R}^{10}$, as well as the root translation $\{ \tau^{(t)} \}_{t=0}^T$ and orientation $\{ \Gamma^{(t)} \}_{t=0}^T$ in the world coordinate system.

\subsection*{Synthetic data generation}

While the kinematic tree of the SMPL model is artist-defined and optimized for rendering, it lacks biomechanical validity: joint centers are inaccurately located, and their degrees of freedom do not reflect true anatomical constraints.

Predicting full-body kinematics that are relevant to clinicians or sport practitioners from a single wearable sensor requires the generation of pairs of anatomically plausible motion and corresponding inertial data.

We first converted BSM—a high-fidelity, full-body musculoskeletal model~\cite{keller2023skel}—from NimblePhysics to OpenSim~\cite{seth2018}, whose joint formalism differs slightly. The model not only incorporates realistic forearm, spine, and shoulder kinematics, but also possesses a set of 105 markers with direct correspondence to vertices on the SMPL mesh surface. As a result, BSM skeletons can be conveniently fitted to each SMPL motion sequence using inverse kinematics to get anatomically sound full-body kinematics from in-the-wild monocular videos. From the resulting trajectories of states, synthetic accelerometer and gyroscope data were generated by simulating the output of a wearable sensor artificially placed on the model at the wrist using OpenSense utilities.

\subsection*{Full-body 3D kinematics prediction}

Given our synthetically generated training dataset $ \mathcal{D} = {(\mathbf{x}^{i}, \mathbf{y}^{i} )}_{i=1}^N $ of N pairs, our task is to learn a function $f : \mathbf{x} \to \mathbf{y}$ that maps inertial data $\mathbf{x}$ to BSM poses $\mathbf{y}$. We adopted Zhou's 6D continuous rotation representation~\cite{Zhou_2019_CVPR} for its superiority in neural network regression tasks. Thus, inputs to the model were sequences of sensor features $\mathbf{x}^{1:L} \in \mathbb{R}^{L \times 9}$ (6D local angular velocity and 3D global acceleration), from which we aimed to predict the corresponding skeleton poses $\mathbf{y}^{1:L} \in \mathbb{R}^{L \times 156}$
(6D orientation in the global coordinate system of BSM's 26 rigid bodies). We set the sequence length $L$ to 32, approximating the average duration of a golf swing from address to impact.

We used the lightweight multilayer perceptron architecture of~\cite{Du_2023_CVPR}, without the conditional diffusion model, for our regression task. The model comprised only two blocks each made of a 1D convolutional layer with kernel size 1 and a fully connected layer (respectively merging temporal and spatial information), together with Layer Norm as pre-normalization, skip connections, and SiLU activation layers. In addition, the model included a first fully connected layer projecting the input data to a 256D latent space, and a final one mapping the latent space back to the output space.

The model was trained on 80\% of the synthetic data for 500 epochs, with a batch size of 512 and the Adam optimizer; the learning rate was initially set to $\num{3e-4}$, before dropping to $\num{3e-5}$ after 225,000 iterations. The weight decay was set to $\num{1e-4}$ for the entire training. The loss function was the mean squared error between the predicted and ground truth 6D rotations. During inference, the model was applied in an autoregressive manner (with no overlap over consecutive batches). We used OpenSim's orientation-based inverse kinematics to pose the BSM model given the predicted rigid body 6D orientations in each frame and obtain instantaneous joint angles and 3D positions.

We evaluated the performance of the network using the mean per joint position error (MPJPE, in cm) and mean per joint rotation error (MPJRE, in degrees) metrics, which are commonly used in the literature to assess the quality of 3D pose reconstruction. The MPJPE is the average Euclidean distance between the predicted and ground truth joint positions, while the MPJRE is the average error between the predicted and ground truth joint angles.

\subsection*{Event detection}

For a fair comparison with the results reported in~\cite{McNally_2019_CVPR_Workshops}, we trained the same shallow RNN decoder to predict golf swing events on all four data splits. The model is a bidirectional LSTM with a single layer of 256 hidden units, trained using the Adam optimizer with a learning rate of $10^{-3}$, a batch size of 512, and for 250 epochs, with early stopping if no improvement in validation loss was observed for 10 consecutive epochs. To account for the high event imbalance ($\approx$16:1 in the real-time videos), we employed weighted cross-entropy loss, with class weights set as the inverse of class frequencies. During training, the model is fed random crops of 32 samples of predicted BSM poses, which also serves as a simple data augmentation in the time domain. At test time, the entire time series is passed to the network in batches of 32 samples, with the last batch zero-padded if shorter than the batch length. The model outputs a vector of length 8, with each element representing the probability of a corresponding event occurring. Following the methodology of the original paper, we evaluated the model's performance using the “Percentage of Correct Events” (PCE) metric. An event is considered correctly predicted if its predicted frame lies within $\pm1$ frame of the ground-truth event frame. PCE is then computed as the percentage of correctly predicted events out of all ground-truth events.

\subsection*{Golf swing tokenization}

We developed a discrete vocabulary of professional golf swing motion primitives by mapping swing sequences to a compressed, discrete latent space. This representation enhances interpretability, simplifies motion editing capabilities, and pairs well with large language models for multimodal applications.

\subsubsection*{VQ-VAE architecture}

Our approach builds upon the Vector Quantized Variational Autoencoder (VQ-VAE;~\cite{van2017neural}), which comprises an encoder that maps observations to a sequence of discrete tokens and a decoder that reconstructs the original sequence from these discrete representations. Specifically, we made slight architectural changes to ACTOR's~\cite{petrovich2021action} Transformer-based motion encoder and decoder: we removed action conditioning and the bias token to make the model class-agnostic, and replaced the learnable distribution tokens with two linear layers that project into the discrete embedding space. Crucially, we treat individual body parts as five distinct components—left arm, right arm, left leg, right leg, and backbone (pelvis + spine + head)—and learn a separate codebook for each. This compositional approach, inspired from recent works in motion modeling~\cite{yi2023generating,zou2024parco}, enables the model to learn more fine-grained and expressive representations for each body part, capturing the nuanced biomechanics of golf swings.

\subsubsection*{Finite Scalar Quantization}

To generate discrete tokens, we use Finite Scalar Quantization (FSQ) due to its simplicity and advantages over traditional vector quantization~\cite{mentzer2023finite}. FSQ obviates the need for auxiliary losses, provides more stable training, and achieves very high codebook utilization with much fewer parameters and none of the common tricks (e.g., codebook reset, EMA on codebook), without sacrificing performance.

In FSQ, the continuous latent representation is projected to a low-dimensional space (typically 3–6 dimensions) where each scalar value is quantized via a bounded mapping operation, $z_i \mapsto \lfloor \frac{L}{2} \rfloor \tanh(z_i)$, followed by rounding to the nearest integer. This produces a quantized vector $\hat{\mathbf{z}}$ corresponding to the nearest point in a fixed grid partition, effectively defining an implicit codebook with precisely controlled size. For a vector $\mathbf{z}$ with $d$ channels, where each entry $z_i$ is mapped to $L$ values, the quantized output $\hat{\mathbf{z}}$ represents one of $\prod_{i=1}^{d} L_i$ unique vectors. In our implementation, we used a 3-dimensional latent space with quantization levels $[7, 6, 5]$, yielding a codebook size of $|C| = 7 \times 6 \times 5 = 210$ unique tokens per body part. To preserve end-to-end differentiability, gradients are propagated through the rounding operation using the straight-through estimator~\cite{bengio2013estimating}, which encourages the model to distribute information across multiple quantization bins.

\subsubsection*{Encoding and decoding process}

Given an input golf swing motion sequence $\mathbf{M}_{1:T} \in \mathbb{R}^{T \times 156}$ with $T$ frames, we first decompose it into five body components:

\begin{align}
    \mathbf{M}_{1:T}^{la} & \in \mathbb{R}^{T \times 36} \quad \text{(left arm)}  \\
    \mathbf{M}_{1:T}^{ra} & \in \mathbb{R}^{T \times 36} \quad \text{(right arm)} \\
    \mathbf{M}_{1:T}^{ll} & \in \mathbb{R}^{T \times 30} \quad \text{(left leg)}  \\
    \mathbf{M}_{1:T}^{rl} & \in \mathbb{R}^{T \times 30} \quad \text{(right leg)} \\
    \mathbf{M}_{1:T}^{b}  & \in \mathbb{R}^{T \times 24} \quad \text{(backbone)}
\end{align}

These components are independently encoded and quantized using FSQ into discrete latent ``motion'' tokens $\mathbf{z} \in \mathbb{Z}^{T \times 5}$, where each row represents the tokens corresponding to the five body parts at a given timestep. Let $\phi$ denote the encoder function and $q$ the FSQ quantization operation. Then, the resulting tokens $\mathbf{z}$ are computed as:

\begin{equation}
    \mathbf{z} = q(\phi(\mathbf{M}_{1:T}))
\end{equation}

The discretized representation is then fed to the decoder $\psi$ to reconstruct the original motion:

\begin{equation}
    \hat{\mathbf{M}}_{1:T} = \psi(\mathbf{z})
\end{equation}

\subsubsection*{Training objective}

We train our VQ-VAE by minimizing a composite loss function that includes both position and velocity reconstruction terms:

\begin{equation}
    \begin{aligned}
        \mathcal{L}_{\text{recon}} = {} & \mathcal{L}_{\text{smooth-}\ell_1}(\mathbf{M}_{1:T}, \mathbf{\hat{M}}_{1:T})                 \\
                                        & + \lambda \cdot \mathcal{L}_{\text{smooth-}\ell_1}(\mathbf{V}_{1:T}, \mathbf{\hat{V}}_{1:T})
    \end{aligned}
\end{equation}

where $\mathbf{V}_{1:T}$ and $\hat{\mathbf{V}}_{1:T}$ represent the velocities of the original and reconstructed motions, respectively. We simply define the velocity of a motion sequence $\mathbf{M}$ as $\mathbf{V}(\mathbf{M}) = [\mathbf{v}_1, \mathbf{v}_2, \ldots, \mathbf{v}_{T-1}]$, where $\mathbf{v}_t = \mathbf{m}_{t+1} - \mathbf{m}_t$ represents the frame-to-frame displacement. The velocity term acts as a regularizer that enhances the quality and temporal smoothness of the reconstructed motions. In our experiments, we set $\lambda = 0.5$ to balance position accuracy and motion smoothness.

\subsection*{Golf swing prior}

\begin{figure*}[t]
    \centering
    \includegraphics[width=\textwidth]{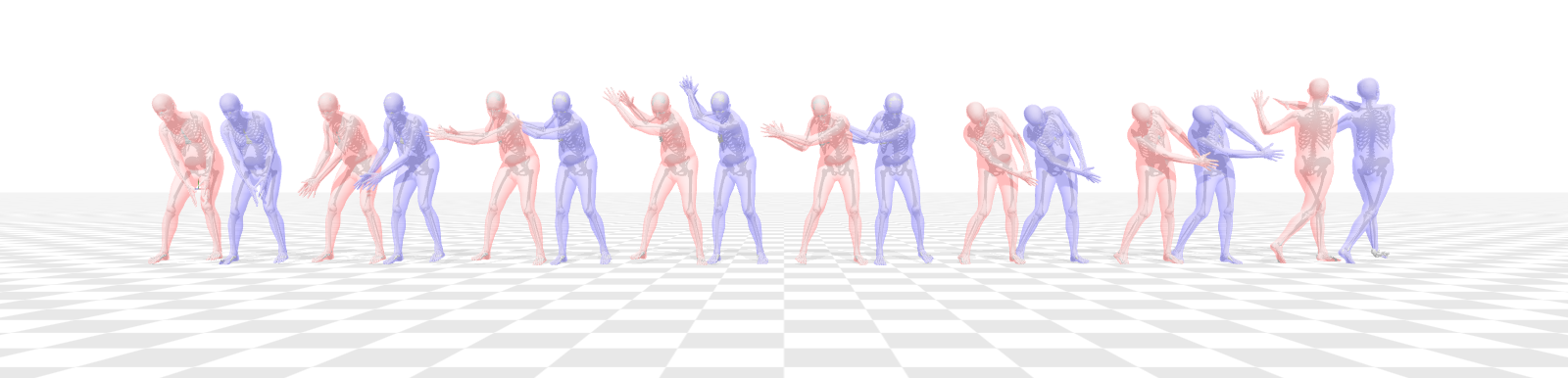}
    \caption{{\bf Qualitative comparison of ground truth and predicted golf swing motion.}
        Visualization of eight key events in a golf swing, showing ground truth (red) and predictions from a wrist-worn wearable (blue) side by side. Events are displayed in sequence: (1) address, (2) toe-up, (3) mid-backswing, (4) top, (5) mid-downswing, (6) impact, (7) mid-follow-through, and (8) finish. The SKEL parametric model~\cite{keller2023skel} and the aitviewer~\cite{Kaufmann_Vechev_aitviewer_2022} were used for 3D mesh rendering. This figure represents one randomly selected sequence from the test set (YouTube ID: 1a2QrnvBUuI).}
    \label{fig:screenshot}
\end{figure*}

\begin{table*}[b]
    \bigskip
    \centering
    \caption{Event-wise and overall Percent of Correct Events (PCE). SwingNet-160 is the MobileNet-based architecture used in GolfDB~\cite{McNally_2019_CVPR_Workshops} A: address; TU: toe-up; MB: mid-backswing; T: top; MD: mid-downswing; I: impact; MFT: mid-follow-through; F: finish.}
    \begin{tabular}{c|cccccccc|c}
        \toprule
                                & A             & TU            & MB            & T             & MD            & I             & MFT           & F             & PCE           \\
        \midrule
        SwingNet-160 (averaged) & 38.7          & \textbf{87.2} & \textbf{92.1} & 90.8          & \textbf{98.3} & 98.4          & 97.2          & 30.7          & 79.2          \\
        \midrule
        Ours (averaged)         & \textbf{49.2} & 85.7          & 91.8          & \textbf{91.6} & 98.2          & \textbf{99.1} & \textbf{97.8} & \textbf{43.6} & \textbf{82.1} \\
        \midrule
        Ours (split 1)          & 57.5          & 84.2          & 92.7          & 93.9          & 98.4          & 99.0          & 99.0          & 46.3          & 83.9          \\
        Ours (split 2)          & 48.6          & 86.4          & 91.7          & 89.5          & 97.8          & 99.0          & 97.8          & 44.1          & 81.8          \\
        Ours (split 3)          & 51.8          & 88.3          & 93.3          & 93.0          & 98.4          & 99.0          & 97.8          & 45.0          & 83.3          \\
        Ours (split 4)          & 39.0          & 84.0          & 89.6          & 89.9          & 98.1          & 99.4          & 96.8          & 39.0          & 79.5          \\
        \bottomrule
    \end{tabular}
    \label{table:pce}
\end{table*}

After training the VQ-VAE encoder and decoder, we aimed to fit a powerful prior over the latent tokens for motion anomaly detection and inpainting. This reduces the gap between the marginal posterior and prior distributions, ensuring that latent variables sampled at test time closely match those seen by the decoder during training~\cite{razavi2019vqvae2}, resulting in more coherent outputs for both tasks.
Given that individual codes derive their semantic meaning only when considered in combination~\cite{mentzer2023finite}, and in light of the 2D structure of our latent space, it is essential to adopt a framework that preserves the spatiotemporal relationships between limbs. We therefore opted for the method described in MoGenTS~\cite{yuan2024mogents}, and trained a bidirectional transformer to predict randomly masked tokens by attending to tokens along both the spatial and temporal dimensions. The architecture consists of a stack of four transformer layers, each with eight self-attention heads and a feedforward layer dimension of $1024$. The model embeds the discrete latent codes into a $256$D continuous space before processing them through the transformer layers.

We followed the masking procedure from~\cite{yuan2024mogents}, sequentially masking (i.e., replacing with a special \texttt{MASK} token) all limbs in a random number of frames, followed by masking individual limbs at random in the remaining unmasked frames. Mask ratios for both dimensions obey the same cosine schedule (inspired from MaskGIT), given by the following concave function: \begin{equation}
    \gamma(\tau) = \cos\left(\frac{\pi\tau}{2}\right),
\end{equation}
where $\tau \sim \mathcal{U}(0, 1)$ is uniformly sampled. Thus, for a sequence of $T$ frames and $N$ limbs, $\gamma(\tau) \cdot T \cdot N$ tokens are masked temporally, and $\gamma(\tau) \cdot N$ limbs are masked spatially in each unmasked frame.
Note that, similar to BERT pretraining~\cite{devlin-etal-2019-bert}, a token selected for masking is only effectively masked $80\%$ of the time, replaced with a random token ($10\%$), or left unchanged ($10\%$). This strategy prevents the model from simply copying input patterns and encourages it to learn robust representations of motion sequences.

\subsection*{Downstream tasks}

\subsubsection*{Anomaly detection and biofeedback}

Our learned prior can condition on any subset of unmasked motion tokens due to the masking regime employed during training, making it well-suited for both anomaly detection and motion inpainting. To demonstrate these capabilities, we process new golf swing sequences through the pretrained VQ-VAE motion encoder to obtain sequences of motion tokens, which are then passed through the bidirectional masked transformer. We use the conditional probabilities output by the prior to detect low-likelihood tokens (threshold $p < 0.05$), which correspond to kinematics that deviate significantly from the professional standard. Analogous to the mask-free editing procedure in Muse~\cite{chang2023muse}, we then mask these anomalous tokens and iteratively sample replacements conditioned on the unmasked tokens—effectively inpainting them with codewords that best reflect professional golfers' techniques.

The inpainted token sequence is subsequently decoded to reconstruct the full-body kinematics, allowing for visual comparison between the original input and the ``corrected'' version. This provides an interpretable form of biofeedback, identifying specific movements that deviate from expert technique and replacing them with corresponding exemplars.

To validate the practical utility of our approach, we analyze a longitudinal case study using a YouTube video (ID: jtJeQJzVMLg) featuring a young male golfer who recorded 24 swings over a 1.5-year period, during which his handicap improved from 50 to 2.2 (as accurately documented for each swing in the video).
We test the extent to which technical improvements captured by our method correlate with the reported handicap progression using Pearson's correlation coefficient, and use the average $L^1$ distance between each quantized golf swing and all samples in our database as the scoring metric.

\subsubsection*{Prediction of sex, club type, age, and player name}

To assess the discriminative power of the learned discrete motion tokens, we evaluated performance on four downstream tasks:
\begin{itemize}
    \item \textbf{Sex classification}: binary classification of player sex (male vs. female);
    \item \textbf{Club type classification}: multi-class classification of the golf club being used (driver, fairway, hybrid, iron, wedge);
    \item \textbf{Age regression}: prediction of player age as a continuous variable;
    \item \textbf{Player identification}: multi-class classification to identify individual players ($N=77$, each with $\geq3$ driver swings).
\end{itemize}

We employed two complementary evaluation strategies: (i) linear probes—i.e., a single projection layer without an activation function—with frozen encoder to assess the linear separability of the learned representations; and (ii) full fine-tuning with a shallow MLP head (two layers with a hidden dimension of $512$, batch normalization, ReLU activation, and a dropout rate of $0.3$).

All models were trained using the Adam optimizer with separate learning rates for the head ($5 \times 10^{-4}$) and backbone ($5 \times 10^{-5}$), a weight decay of $1 \times 10^{-5}$, and early stopping based on validation performance. Classification tasks used cross-entropy loss and accuracy as the evaluation metric, while age regression used $\ell_1$ loss and mean absolute error.
We used a 70/15/15 random split for training, validation, and testing across all tasks. In the player identification setting, we applied a stratified split to ensure balanced representation across individuals. For age regression, we used a player-disjoint split, evaluating the model on individuals entirely unseen during training to assess generalization to new subjects.

To interpret which aspects of a swing were most informative for identifying individual golfers, we employed Layer-wise Relevance Propagation (LRP; ~\cite{bach2015pixel})—an established explainability method for nonlinear models, previously applied in gait biomechanics (e.g.,~\cite{horst2019explaining})—using the high-level Python framework Zennit~\cite{anders2021software}. LRP decomposes the output of a classifier into token-level relevance scores, assigning a value to each token that reflects its contribution to the model's decision. Positive scores indicate features that support the classification, while negative scores reflect features that argue against it. By applying LRP to our trained player identification model, we obtained relevance maps across both time and body segments, allowing us to localize the biomechanical elements most distinctive to each player. This approach enables a principled interpretation of golf swing signatures, grounded in the model's learned representations.

\section*{Results}

We evaluated our system's ability to reconstruct full-body golf kinematics from a single wrist-worn inertial sensor, detect key swing events, and capture distinctive movement signatures across individuals and conditions.

\begin{figure*}[!htbp]
    \centering
    \includegraphics[width=0.5\textwidth]{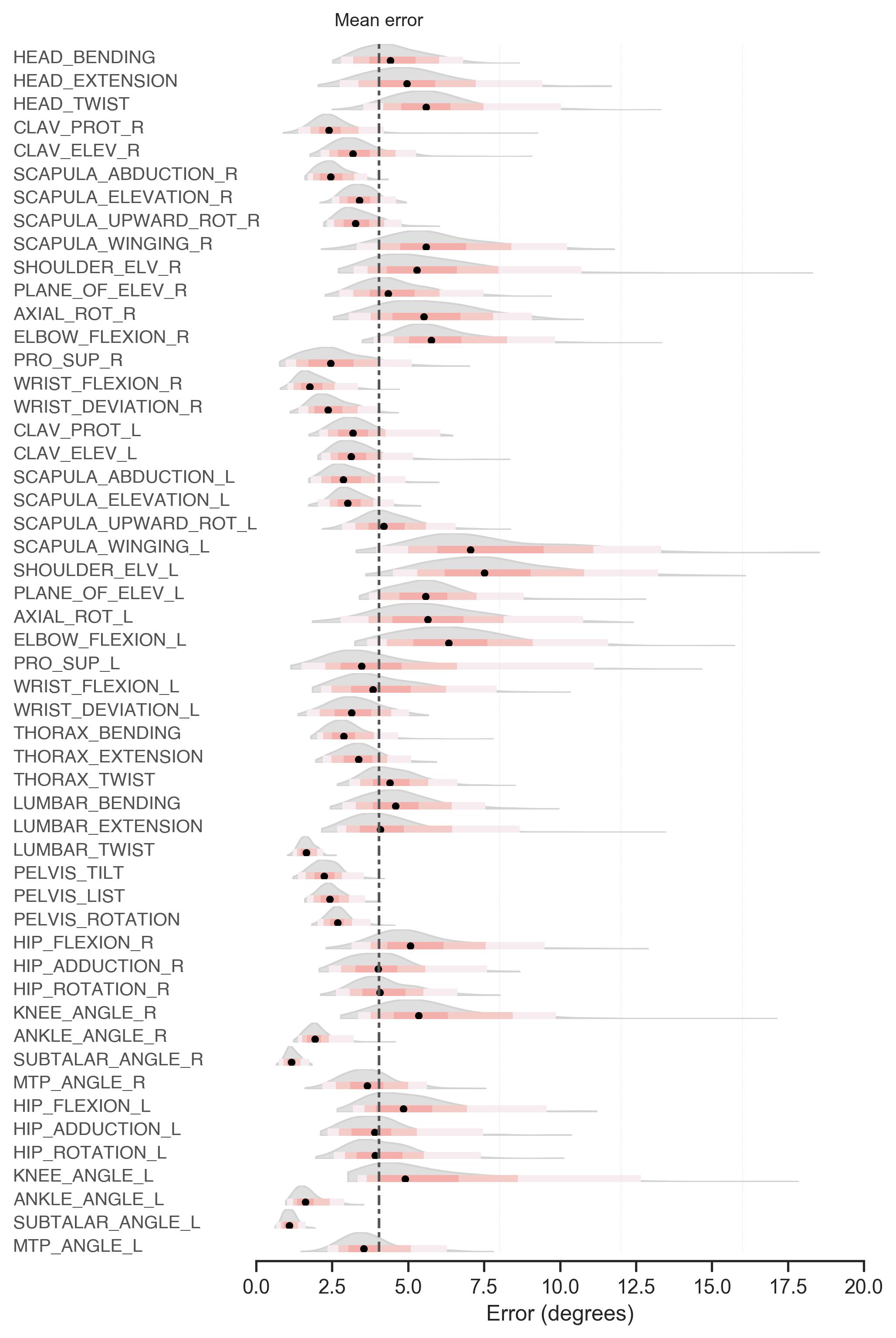}
    \caption{{\bf Ridgeline plot of errors in joint angle estimates.} The plot shows the distribution of errors for each of the 52 degrees of freedom of the BSM biomechanical model. The average reconstruction error is $4.0\pm2.1\degree$. In addition to kernel density estimates (in gray) are median errors (black dots), interquartile ranges (salmon), and 80\% (pink) and 95\% (light pink) confidence intervals.}
    \label{fig:errors}
\end{figure*}

Figure~\ref{fig:screenshot} illustrates the qualitative agreement between predicted and ground-truth joint positions across eight key phases of the golf swing. Quantitatively, our model estimated three-dimensional body joint positions from a single wrist-worn inertial sensor with a mean error of $5.3\pm1.1$ cm and joint rotations within $4.0\pm2.1\degree$ (Fig.~\ref{fig:errors}). Notably, wrist trajectories were reconstructed with an average error of $6.4\pm1.2$ cm.

Using the reconstructed full-body motion, our approach outperformed prior methods in detecting swing events (Table~\ref{table:pce}), achieving a PCE of 82.1\% across eight annotated events. Compared to SwingNet~\cite{McNally_2019_CVPR_Workshops}, our model improved accuracy by three percentage points while using only 1/7th of the parameters ($\num{0.78e6}$ vs. $\num{5.38e6}$). Improvements were especially notable for detecting the address and finish events, which are often challenging due to low motion salience. Overall, six out of eight swing events were detected with an average accuracy of 94.0\%.

\begin{figure}[b]
    \centering
    \includegraphics[width=0.48\textwidth]{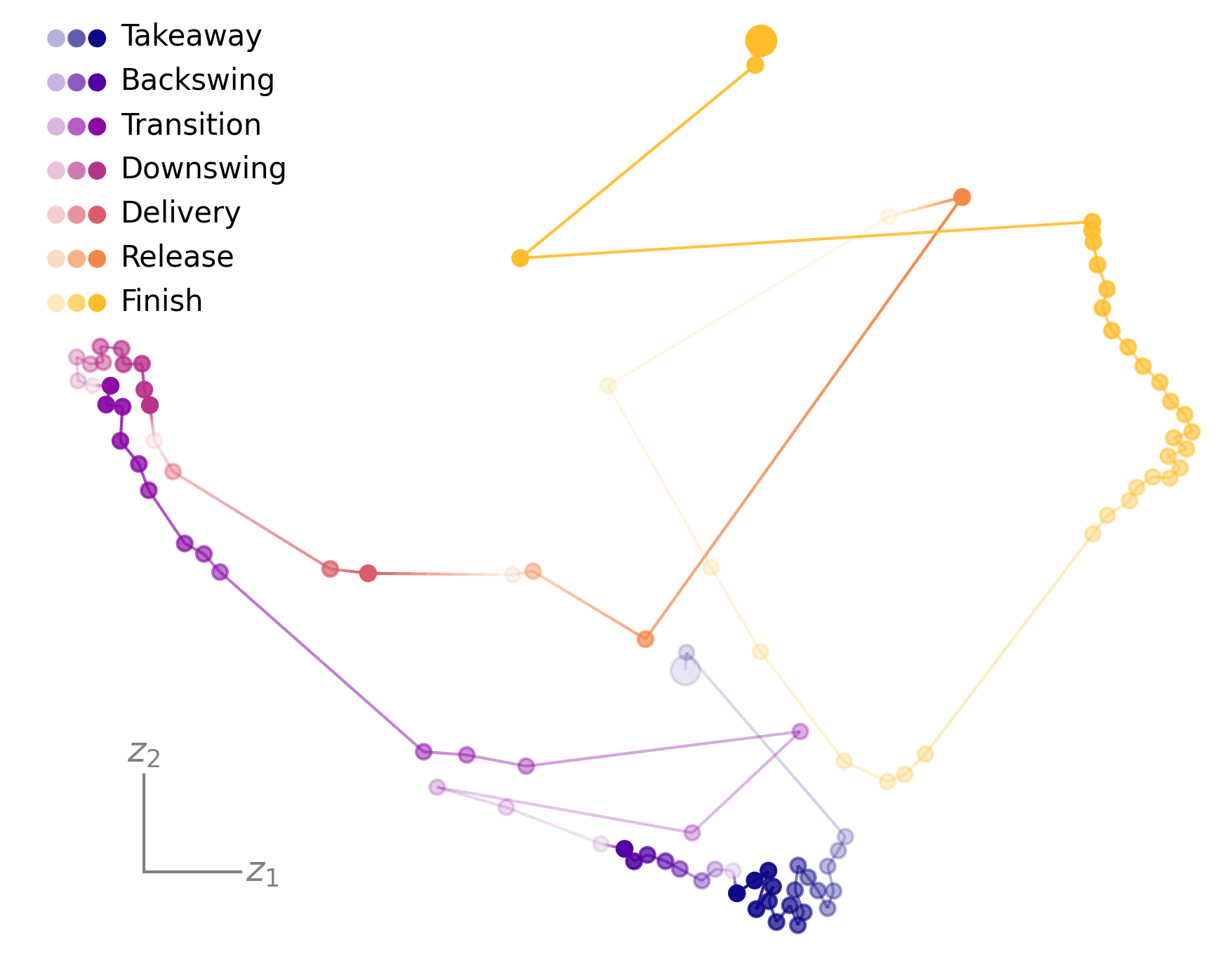}
    \caption{{\bf t-SNE visualization of the latent space of motion tokens.} The tokens are color-coded by the golf swing phase they represent, with opacity increasing as the swing progresses. Each colored dot represents a unique instantaneous pose.}
    \label{fig:tsne}
\end{figure}

\begin{figure*}[!htbp]
    \centering
    \includegraphics[width=\textwidth]{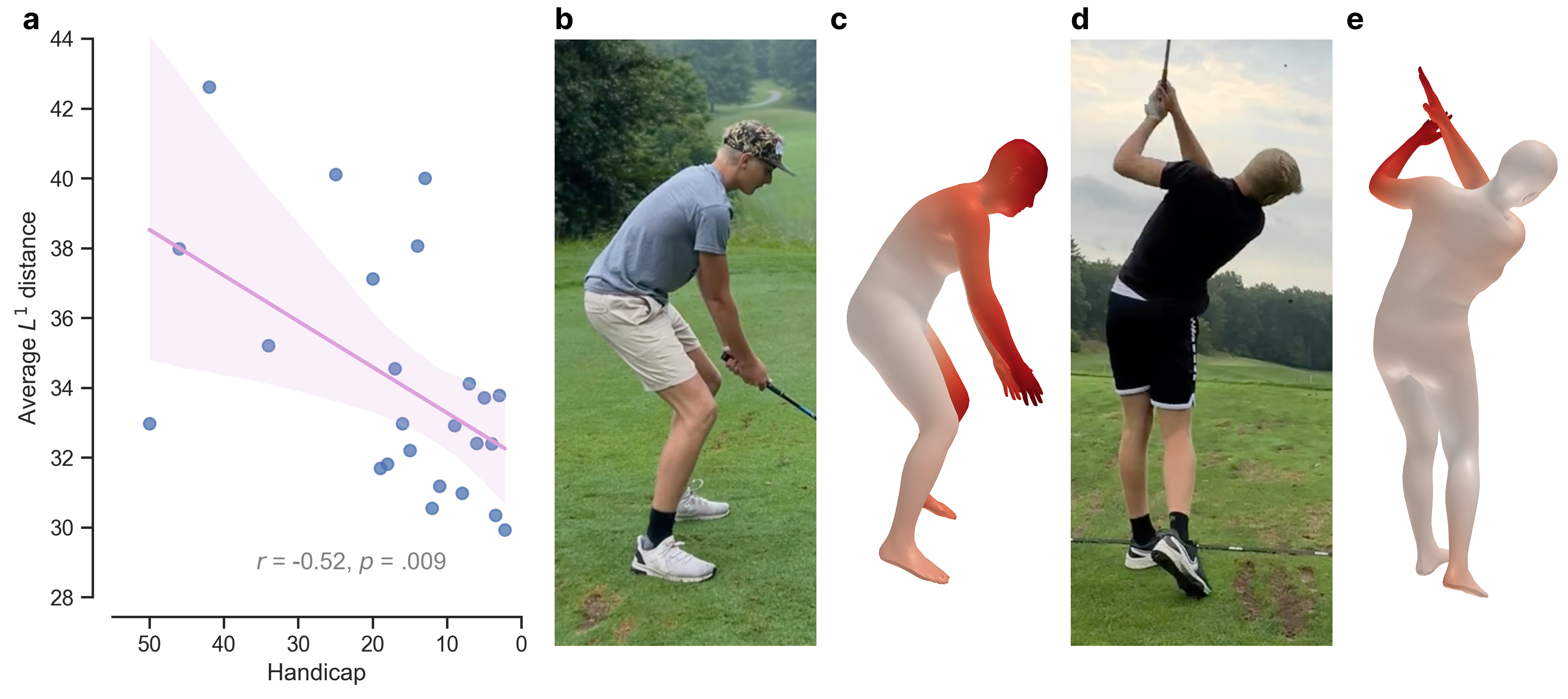}
    \caption{{\bf Longitudinal case study of handicap improvement.} (a) The straight plum line represents the best linear fit while the light pink band corresponds to the 95\% confidence interval for the regression slope estimated using bootstrapping ($N=1000$ trials). As the player's handicap improved from 50 to 2.2, the $L^1$ distance between the quantized swings ($N=24$) and our professional database decreased from 33.0 to 29.9. (b, c) Video frame and corresponding 3D reconstruction showing the instant of maximum error (based on Euclidean distance between original and inpainted sequences) for a high-handicap (50) swing. The primary technical flaw is an excessively bent posture with arms positioned too close to the body. (d, e) Equivalent comparison for a low-handicap (3) swing. In this case, the main error occurs during follow-through, with restricted arm movement and incomplete extension of the right arm. In both reconstructions (c, e), red coloration intensity indicates the magnitude of positional error.}
    \label{fig:progress}
\end{figure*}

\begin{figure}[!htbp]
    \centering
    \includegraphics[width=0.48\textwidth]{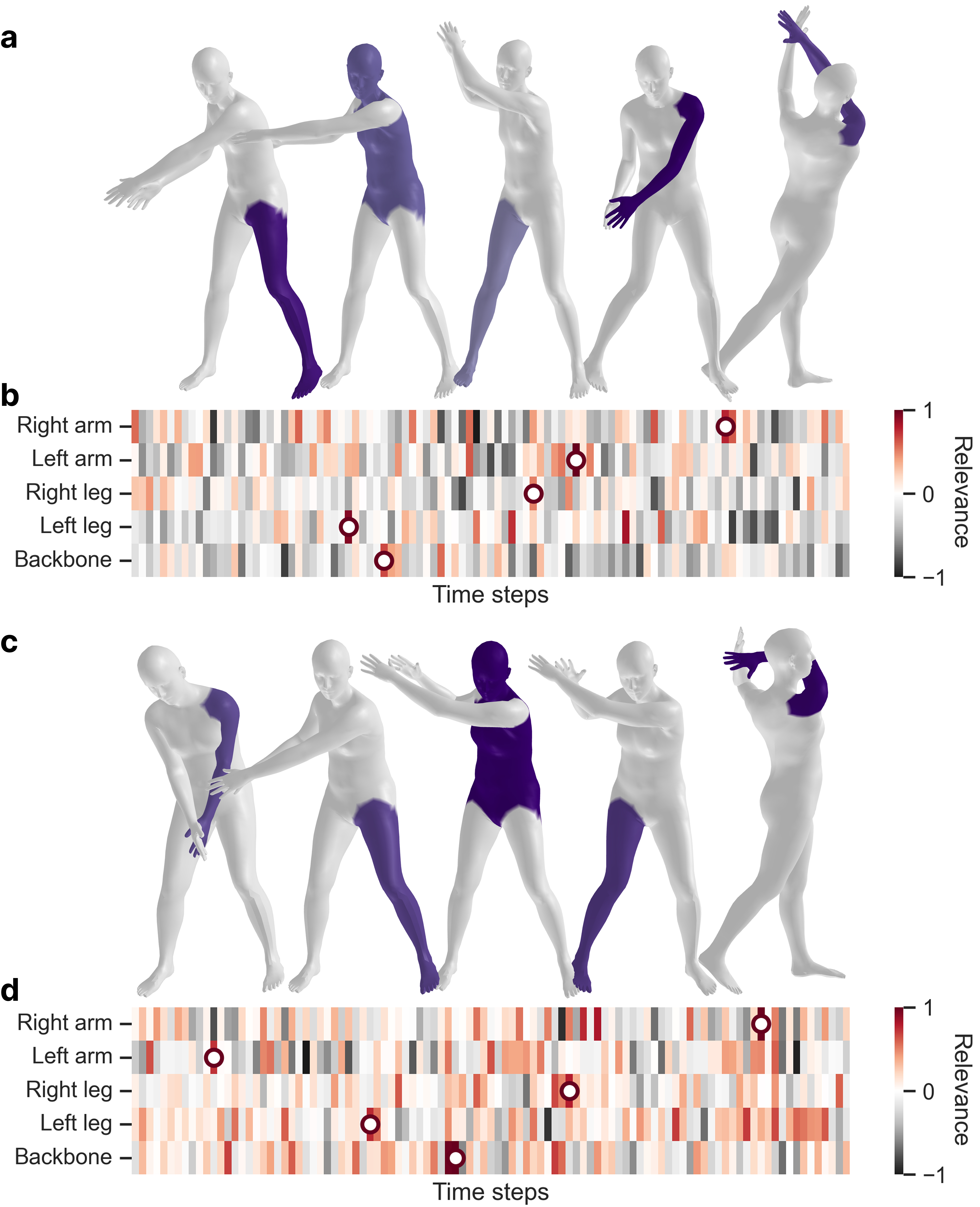}
    \caption{{\bf Swing signatures of two different players.} (a, c) 3D reconstructions at the five time points most relevant to player identity classification, as determined by layer-wise relevance propagation. (b, d) Token relevance heatmaps: tokens supporting the classification appear in red, contradictory inputs in black, and neutral tokens in white. Red dots mark the points of maximum relevance.}
    \label{fig:signature}
\end{figure}

While quantization is inherently lossy, our VQ-VAE tokenization maintained high fidelity, achieving an MPJPE of only $2.9\pm2.0$ cm. Ablation revealed the critical role of our compositional tokenization approach: using a single codebook for the entire body increased the error to $7.2\pm6.2$ cm. Codebook utilization was remarkably high, exceeding $96\%$ across body parts. The learned token space (Fig.~\ref{fig:tsne}) exhibited distinct clusters and loops corresponding to the different swing phases.
The golf swing prior model reached a masked token prediction accuracy of 80.9\% on the test set, and yielded an MPJPE of $2.1\pm0.2$ cm and an MPJRE of $0.9\pm0.1\degree$ upon decoding, indicating strong predictive capacity for professional golf motions.

In a longitudinal case study, we found a statistically significant moderate negative correlation between the average $L^1$ distance to the professional database and the player's handicap as it improved from 50 to 2.2 ($r = -0.52$, $p = .009$; Fig.~\ref{fig:progress}).

The discriminative power of our learned representations is summarized in Table~\ref{table:downstream}. Fine-tuned models achieved accuracies of 97.6\% for sex classification, 80.3\% for club type classification, and 87.1\% for player identification. Age regression on held-out golfers yielded a mean absolute error of 4.70 years.
Even with linear probes, performance remained strong, suggesting a well-structured and semantically meaningful latent space—though identifying individuals proved more challenging without fine-tuning, with accuracy dropping to $72.6\%$.

The relevance of individual motion tokens in accurately predicting a player's identity varied greatly among players and body parts (Fig.~\ref{fig:signature}). Furthermore, token relevance scores over time were highly unequal: subtle movements throughout the entire swing—rather than specific swing phases—were identified as unique to each golfer. Notably, fewer tokens were highly informative for classifying the first player compared to the second (Fig.~\ref{fig:signature}b,d).

\medskip
\begin{table}[!htbp]
    \centering
    \caption{{\bf Classification and regression performance using learned motion representations.} Performance metrics for different downstream tasks using both linear probes (frozen encoder) and full fine-tuning approaches. Classification tasks (sex, club type, and player identification) report accuracy, while age regression reports MAE in years.}
    \begin{tabular}{lcc}
        \toprule
        Task                     & Linear probe & Fine-tuning \\
        \midrule
        Sex classification       & 0.943        & 0.976       \\
        Club type classification & 0.754        & 0.803       \\
        Player identification    & 0.726        & 0.871       \\
        Age regression           & 6.55         & 4.70        \\
        \bottomrule
    \end{tabular}
    \label{table:downstream}
\end{table}

\section*{Discussion}

This work introduces a novel and practical framework for capturing, analyzing, and interpreting full-body golf swing kinematics using a single wrist-worn sensor. By integrating state-of-the-art 3D motion reconstruction with powerful, discrete representations, we build a large-scale dataset of professional golf swing primitives and offer a scalable, accessible, and field-based solution for golf swing analysis. The resulting system allows interpretable, personalized, and data-driven insights into swing biomechanics, individual playing styles, and training progression.

Using only a single wrist-worn sensor, our system estimated wrist trajectories within 6.4 cm—substantially outperforming recent benchmarks~\cite{kim2024enhancing}. More importantly, the precision of our kinematic estimates (under 5$\degree$ of joint angle error on average) enables accurate full-body motion reconstruction using an OpenSim musculoskeletal model compliant with International Society of Biomechanics standards. This ensures anatomical realism and enhances reproducibility across studies, paving the way for standardized reporting of swing kinematics alongside existing guidelines for epidemiological data on golf injury~\cite{murray2020international}.
Crucially, by delivering high-fidelity, anatomically accurate motion capture in the field, our approach bridges the longstanding gap between controlled lab-based assessments and golf performance in natural settings. This opens new possibilities for ecologically valid, large-scale cohort studies of golf biomechanics, including prospective investigations of injury risk.
Such biomechanical fidelity—particularly at the spine, shoulder, and knee—may be especially valuable given their high susceptibility to injury in golfers~\cite{robinson2019systematic,robinson2024prospective}. Precise motion tracking could aid in early detection of harmful movement patterns preceding clinical symptoms and inform targeted, scalable interventions for injury prevention and rehabilitation without the constraints of laboratory infrastructure.

Beyond technical accuracy, our results highlight the individuality embedded in golf swing mechanics. The system identified players with 87.1\% accuracy using tokenized motion alone, demonstrating that each golfer exhibits subtle, consistent, and highly person-specific movement signatures. Notably, cues driving the classification decision were distributed across all body segments and swing phases rather than concentrated in a single salient feature, suggesting that individual identification emerges from integrated whole-body coordination. These findings reinforce a growing view that individual variation represents not merely noise or error but a defining characteristic of skilled movement~\cite{glazier2024ecological}. Recognizing and respecting this variability, instead of enforcing convergence toward a single ``ideal'' swing, may help reframe foundational assumptions in golf instruction and performance evaluation.

Contrary to the common belief that swings are invariant across clubs, our model predicted club type with 80.3\% accuracy, reflecting slight yet systematic technical adaptations to equipment. These findings support prior laboratory studies of low-handicap players~\cite{egret2003analysis,joyce2013three} and demonstrate that such distinctions persist even on-course among professional golfers. Additionally, motion tokens captured biologically meaningful variance: sex classification reached 97.6\% accuracy, and age was predicted with a mean absolute error of 4.70 years on unseen golfers. These results suggest that tokenized motion encodes latent biomechanical biomarkers shaped by both intrinsic factors and task-specific constraints—opening avenues for movement-based phenotyping, differentiated club design, and more nuanced investigations of motor aging and skill development.

Our longitudinal analysis demonstrates the potential for feedback-driven training. As a player's handicap improved, their swing became measurably more similar to professional standards. This suggests that motion tokens, in combination with a large ``vocabulary'' of professional motion primitives, provide a means to monitor and quantify technical progression—making objective, evidence-based feedback more accessible and actionable, especially for recreational players seeking to track development and stay motivated.

Despite its strengths, our approach presents several limitations that warrant further investigation.
First, our models were trained on synthetic inertial data derived from video-based reconstructions, and the generalizability to real-world sensor inputs—subject to noise, calibration errors, and drift—remains an open question. Generative approaches, such as diffusion models (e.g.,~\cite{Van_Wouwe_2024_CVPR}), may offer enhanced robustness in these settings.
Second, our pipeline relies on WHAM~\cite{Shin_2024_CVPR} for video-based motion reconstruction, which, despite offering state-of-the-art performance, only provides pseudo-ground truth. Additionally, hand articulation is not modeled, limiting fine-grained analyses of grip biomechanics. Further validation against gold-standard optical motion capture is warranted.
Third, the current system does not track the club head, precluding direct measurement of key performance variables such as clubhead speed, swing plane, and face angle. Incorporating club motion could not only enhance performance analysis, but also improve full-body pose estimation by providing additional kinematic constraints.
Last, while WHAM is designed for real-time use and an implementation on mobile hardware is considered feasible, our pipeline does not yet run natively on-device; achieving real-time feedback will require further development.

\section*{Acknowledgments}

I warmly thank Erwan Delhaye, Maxime Vidal, Shaokai Ye, Mu Zhou, and Aristotelis Economides for critical feedback.







\section*{Bibliography}

\begin{thebibliography}{10}
    \expandafter\ifx\csname url\endcsname\relax
        \def\url#1{\texttt{#1}}\fi
    \expandafter\ifx\csname urlprefix\endcsname\relax\def\urlprefix{URL }\fi
    \providecommand{\bibinfo}[2]{#2}
    \providecommand{\eprint}[2][]{\url{#2}}

    \bibitem{golfstatsus2024}
    \bibinfo{author}{NGF}.
    \newblock \bibinfo{title}{Golf industry facts 2024}.
    \newblock \bibinfo{howpublished}{\url{https://www.ngf.org/the-clubhouse/golf-industry-research/}}.
    \newblock \bibinfo{note}{Accessed: 2024-05-08}.

    \bibitem{golfstatsra2024}
    \bibinfo{author}{R\&A}.
    \newblock \bibinfo{title}{Global golf participation 2024}.
    \newblock \bibinfo{howpublished}{\url{https://www.ega-golf.ch/sites/ega/files/ckeditor/Sporting\%20Insights\%20Report_The\%20R\%26A_Global\%20Golf\%20Participation\%202023_August\%202024.pdf}}.
    \newblock \bibinfo{note}{Accessed: 2024-05-08}.

    \bibitem{theriault1996golf}
    \bibinfo{author}{Thériault, G.}, \bibinfo{author}{Lacoste, E.}, \bibinfo{author}{Gadoury, M.}, \bibinfo{author}{Ouellet, S.} \& \bibinfo{author}{Leblanc, C.}
    \newblock \bibinfo{title}{Golf injury characteristics: a survey from 528 golfers}.
    \newblock \emph{\bibinfo{journal}{Medicine \& Science in Sports \& Exercise}} \textbf{\bibinfo{volume}{28}}, \bibinfo{pages}{65} (\bibinfo{year}{1996}).

    \bibitem{hume2005role}
    \bibinfo{author}{Hume, P.~A.}, \bibinfo{author}{Keogh, J.} \& \bibinfo{author}{Reid, D.}
    \newblock \bibinfo{title}{The role of biomechanics in maximising distance and accuracy of golf shots}.
    \newblock \emph{\bibinfo{journal}{Sports medicine}} \textbf{\bibinfo{volume}{35}}, \bibinfo{pages}{429--449} (\bibinfo{year}{2005}).

    \bibitem{keogh2012evidence}
    \bibinfo{author}{Keogh, J.~W.} \& \bibinfo{author}{Hume, P.~A.}
    \newblock \bibinfo{title}{Evidence for biomechanics and motor learning research improving golf performance}.
    \newblock \emph{\bibinfo{journal}{Sports Biomechanics}} \textbf{\bibinfo{volume}{11}}, \bibinfo{pages}{288--309} (\bibinfo{year}{2012}).

    \bibitem{theriault1998golf}
    \bibinfo{author}{Thériault, G.} \& \bibinfo{author}{Lachance, P.}
    \newblock \bibinfo{title}{Golf injuries: an overview}.
    \newblock \emph{\bibinfo{journal}{Sports medicine}} \textbf{\bibinfo{volume}{26}}, \bibinfo{pages}{43--57} (\bibinfo{year}{1998}).

    \bibitem{bourgain2022golf}
    \bibinfo{author}{Bourgain, M.}, \bibinfo{author}{Rouch, P.}, \bibinfo{author}{Rouillon, O.}, \bibinfo{author}{Thoreux, P.} \& \bibinfo{author}{Sauret, C.}
    \newblock \bibinfo{title}{Golf swing biomechanics: A systematic review and methodological recommendations for kinematics}.
    \newblock \emph{\bibinfo{journal}{Sports}} \textbf{\bibinfo{volume}{10}}, \bibinfo{pages}{91} (\bibinfo{year}{2022}).

    \bibitem{smith2015golf}
    \bibinfo{author}{Smith, A.}, \bibinfo{author}{Roberts, J.}, \bibinfo{author}{Wallace, E.}, \bibinfo{author}{Kong, P.~W.} \& \bibinfo{author}{Forrester, S.}
    \newblock \bibinfo{title}{Golf coaches' perceptions of key technical swing parameters compared to biomechanical literature}.
    \newblock \emph{\bibinfo{journal}{International Journal of Sports Science \& Coaching}} \textbf{\bibinfo{volume}{10}}, \bibinfo{pages}{739--755} (\bibinfo{year}{2015}).

    \bibitem{gloersen2018technique}
    \bibinfo{author}{Gl{\o}ersen, {\O}.}, \bibinfo{author}{Myklebust, H.}, \bibinfo{author}{Hall{\'e}n, J.} \& \bibinfo{author}{Federolf, P.}
    \newblock \bibinfo{title}{Technique analysis in elite athletes using principal component analysis}.
    \newblock \emph{\bibinfo{journal}{Journal of sports sciences}} \textbf{\bibinfo{volume}{36}}, \bibinfo{pages}{229--237} (\bibinfo{year}{2018}).

    \bibitem{suo2024motion}
    \bibinfo{author}{Suo, X.}, \bibinfo{author}{Tang, W.} \& \bibinfo{author}{Li, Z.}
    \newblock \bibinfo{title}{Motion capture technology in sports scenarios: a survey}.
    \newblock \emph{\bibinfo{journal}{Sensors}} \textbf{\bibinfo{volume}{24}}, \bibinfo{pages}{2947} (\bibinfo{year}{2024}).

    \bibitem{yamamoto2023extracting}
    \bibinfo{author}{Yamamoto, K.} \emph{et~al.}
    \newblock \bibinfo{title}{Extracting proficiency differences and individual characteristics in golfers' swing using single-video markerless motion analysis}.
    \newblock \emph{\bibinfo{journal}{Frontiers in Sports and Active Living}} \textbf{\bibinfo{volume}{5}}, \bibinfo{pages}{1272038} (\bibinfo{year}{2023}).

    \bibitem{jiang2022golfpose}
    \bibinfo{author}{Jiang, Z.}, \bibinfo{author}{Ji, H.}, \bibinfo{author}{Menaker, S.} \& \bibinfo{author}{Hwang, J.-N.}
    \newblock \bibinfo{title}{Golfpose: Golf swing analyses with a monocular camera based human pose estimation}.
    \newblock In \emph{\bibinfo{booktitle}{2022 IEEE International Conference on Multimedia and Expo Workshops (ICMEW)}}, \bibinfo{pages}{1--6} (\bibinfo{organization}{IEEE}, \bibinfo{year}{2022}).

    \bibitem{liao2022ai}
    \bibinfo{author}{Liao, C.-C.}, \bibinfo{author}{Hwang, D.-H.} \& \bibinfo{author}{Koike, H.}
    \newblock \bibinfo{title}{Ai golf: Golf swing analysis tool for self-training}.
    \newblock \emph{\bibinfo{journal}{IEEE Access}} \textbf{\bibinfo{volume}{10}}, \bibinfo{pages}{106286--106295} (\bibinfo{year}{2022}).

    \bibitem{ju2023golfmate}
    \bibinfo{author}{Ju, C.-Y.}, \bibinfo{author}{Kim, J.-H.} \& \bibinfo{author}{Lee, D.-H.}
    \newblock \bibinfo{title}{Golfmate: Enhanced golf swing analysis tool through pose refinement network and explainable golf swing embedding for self-training}.
    \newblock \emph{\bibinfo{journal}{Applied Sciences}} \textbf{\bibinfo{volume}{13}}, \bibinfo{pages}{11227} (\bibinfo{year}{2023}).

    \bibitem{lee2025golfpose}
    \bibinfo{author}{Lee, M.-H.}, \bibinfo{author}{Zhang, Y.-C.}, \bibinfo{author}{Wu, K.-R.} \& \bibinfo{author}{Tseng, Y.-C.}
    \newblock \bibinfo{title}{Golfpose: From regular posture to golf swing posture}.
    \newblock In \emph{\bibinfo{booktitle}{International Conference on Pattern Recognition}}, \bibinfo{pages}{387--402} (\bibinfo{organization}{Springer}, \bibinfo{year}{2025}).

    \bibitem{McNally_2019_CVPR_Workshops}
    \bibinfo{author}{McNally, W.} \emph{et~al.}
    \newblock \bibinfo{title}{Golfdb: A video database for golf swing sequencing}.
    \newblock In \emph{\bibinfo{booktitle}{Proceedings of the IEEE/CVF Conference on Computer Vision and Pattern Recognition (CVPR) Workshops}} (\bibinfo{year}{2019}).

    \bibitem{Shin_2024_CVPR}
    \bibinfo{author}{Shin, S.}, \bibinfo{author}{Kim, J.}, \bibinfo{author}{Halilaj, E.} \& \bibinfo{author}{Black, M.~J.}
    \newblock \bibinfo{title}{Wham: Reconstructing world-grounded humans with accurate 3d motion}.
    \newblock In \emph{\bibinfo{booktitle}{Proceedings of the IEEE/CVF Conference on Computer Vision and Pattern Recognition (CVPR)}}, \bibinfo{pages}{2070--2080} (\bibinfo{year}{2024}).

    \bibitem{Mahmood_2019_ICCV}
    \bibinfo{author}{Mahmood, N.}, \bibinfo{author}{Ghorbani, N.}, \bibinfo{author}{Troje, N.~F.}, \bibinfo{author}{Pons-Moll, G.} \& \bibinfo{author}{Black, M.~J.}
    \newblock \bibinfo{title}{Amass: Archive of motion capture as surface shapes}.
    \newblock In \emph{\bibinfo{booktitle}{Proceedings of the IEEE/CVF International Conference on Computer Vision (ICCV)}}, \bibinfo{pages}{5442--5451} (\bibinfo{year}{2019}).

    \bibitem{loper2015smpl}
    \bibinfo{author}{Loper, M.}, \bibinfo{author}{Mahmood, N.}, \bibinfo{author}{Romero, J.}, \bibinfo{author}{Pons-Moll, G.} \& \bibinfo{author}{Black, M.~J.}
    \newblock \bibinfo{title}{Smpl: a skinned multi-person linear model}.
    \newblock \emph{\bibinfo{journal}{ACM Trans. Graph.}} \textbf{\bibinfo{volume}{34}} (\bibinfo{year}{2015}).
    \newblock \urlprefix\url{https://doi.org/10.1145/2816795.2818013}.

    \bibitem{keller2023skel}
    \bibinfo{author}{Keller, M.} \emph{et~al.}
    \newblock \bibinfo{title}{From skin to skeleton: Towards biomechanically accurate 3d digital humans}.
    \newblock \emph{\bibinfo{journal}{ACM Trans. Graph.}} \textbf{\bibinfo{volume}{42}} (\bibinfo{year}{2023}).
    \newblock \urlprefix\url{https://doi.org/10.1145/3618381}.

    \bibitem{seth2018}
    \bibinfo{author}{Seth, A.} \emph{et~al.}
    \newblock \bibinfo{title}{Opensim: Simulating musculoskeletal dynamics and neuromuscular control to study human and animal movement}.
    \newblock \emph{\bibinfo{journal}{PLOS Computational Biology}} \textbf{\bibinfo{volume}{14}}, \bibinfo{pages}{1--20} (\bibinfo{year}{2018}).
    \newblock \urlprefix\url{https://doi.org/10.1371/journal.pcbi.1006223}.

    \bibitem{Zhou_2019_CVPR}
    \bibinfo{author}{Zhou, Y.}, \bibinfo{author}{Barnes, C.}, \bibinfo{author}{Lu, J.}, \bibinfo{author}{Yang, J.} \& \bibinfo{author}{Li, H.}
    \newblock \bibinfo{title}{On the continuity of rotation representations in neural networks}.
    \newblock In \emph{\bibinfo{booktitle}{Proceedings of the IEEE/CVF Conference on Computer Vision and Pattern Recognition (CVPR)}} (\bibinfo{year}{2019}).

    \bibitem{Du_2023_CVPR}
    \bibinfo{author}{Du, Y.} \emph{et~al.}
    \newblock \bibinfo{title}{Avatars grow legs: Generating smooth human motion from sparse tracking inputs with diffusion model}.
    \newblock In \emph{\bibinfo{booktitle}{Proceedings of the IEEE/CVF Conference on Computer Vision and Pattern Recognition (CVPR)}}, \bibinfo{pages}{481--490} (\bibinfo{year}{2023}).

    \bibitem{van2017neural}
    \bibinfo{author}{Van Den~Oord, A.}, \bibinfo{author}{Vinyals, O.} \emph{et~al.}
    \newblock \bibinfo{title}{Neural discrete representation learning}.
    \newblock \emph{\bibinfo{journal}{Advances in neural information processing systems}} \textbf{\bibinfo{volume}{30}} (\bibinfo{year}{2017}).

    \bibitem{petrovich2021action}
    \bibinfo{author}{Petrovich, M.}, \bibinfo{author}{Black, M.~J.} \& \bibinfo{author}{Varol, G.}
    \newblock \bibinfo{title}{Action-conditioned 3d human motion synthesis with transformer vae}.
    \newblock In \emph{\bibinfo{booktitle}{Proceedings of the IEEE/CVF International Conference on Computer Vision}}, \bibinfo{pages}{10985--10995} (\bibinfo{year}{2021}).

    \bibitem{yi2023generating}
    \bibinfo{author}{Yi, H.} \emph{et~al.}
    \newblock \bibinfo{title}{Generating holistic 3d human motion from speech}.
    \newblock In \emph{\bibinfo{booktitle}{Proceedings of the IEEE/CVF Conference on Computer Vision and Pattern Recognition}}, \bibinfo{pages}{469--480} (\bibinfo{year}{2023}).

    \bibitem{zou2024parco}
    \bibinfo{author}{Zou, Q.} \emph{et~al.}
    \newblock \bibinfo{title}{Parco: Part-coordinating text-to-motion synthesis}.
    \newblock In \emph{\bibinfo{booktitle}{European Conference on Computer Vision}}, \bibinfo{pages}{126--143} (\bibinfo{organization}{Springer}, \bibinfo{year}{2024}).

    \bibitem{mentzer2023finite}
    \bibinfo{author}{Mentzer, F.}, \bibinfo{author}{Minnen, D.}, \bibinfo{author}{Agustsson, E.} \& \bibinfo{author}{Tschannen, M.}
    \newblock \bibinfo{title}{Finite scalar quantization: Vq-vae made simple}.
    \newblock \emph{\bibinfo{journal}{arXiv preprint arXiv:2309.15505}}  (\bibinfo{year}{2023}).

    \bibitem{bengio2013estimating}
    \bibinfo{author}{Bengio, Y.}, \bibinfo{author}{L{\'e}onard, N.} \& \bibinfo{author}{Courville, A.}
    \newblock \bibinfo{title}{Estimating or propagating gradients through stochastic neurons for conditional computation}.
    \newblock \emph{\bibinfo{journal}{arXiv preprint arXiv:1308.3432}}  (\bibinfo{year}{2013}).

    \bibitem{Kaufmann_Vechev_aitviewer_2022}
    \bibinfo{author}{Kaufmann, M.}, \bibinfo{author}{Vechev, V.} \& \bibinfo{author}{Mylonopoulos, D.}
    \newblock \bibinfo{title}{{aitviewer}} (\bibinfo{year}{2022}).
    \newblock \urlprefix\url{https://github.com/eth-ait/aitviewer}.

    \bibitem{razavi2019vqvae2}
    \bibinfo{author}{Razavi, A.}, \bibinfo{author}{Van~den Oord, A.} \& \bibinfo{author}{Vinyals, O.}
    \newblock \bibinfo{title}{Generating diverse high-fidelity images with vq-vae-2}.
    \newblock \emph{\bibinfo{journal}{Advances in neural information processing systems}} \textbf{\bibinfo{volume}{32}} (\bibinfo{year}{2019}).

    \bibitem{yuan2024mogents}
    \bibinfo{author}{Yuan, W.} \emph{et~al.}
    \newblock \bibinfo{title}{Mogents: Motion generation based on spatial-temporal joint modeling}.
    \newblock \emph{\bibinfo{journal}{Advances in Neural Information Processing Systems}} \textbf{\bibinfo{volume}{37}}, \bibinfo{pages}{130739--130763} (\bibinfo{year}{2024}).

    \bibitem{devlin-etal-2019-bert}
    \bibinfo{author}{Devlin, J.}, \bibinfo{author}{Chang, M.-W.}, \bibinfo{author}{Lee, K.} \& \bibinfo{author}{Toutanova, K.}
    \newblock \bibinfo{title}{{BERT}: Pre-training of deep bidirectional transformers for language understanding}.
    \newblock In \bibinfo{editor}{Burstein, J.}, \bibinfo{editor}{Doran, C.} \& \bibinfo{editor}{Solorio, T.} (eds.) \emph{\bibinfo{booktitle}{Proceedings of the 2019 Conference of the North {A}merican Chapter of the Association for Computational Linguistics: Human Language Technologies, Volume 1 (Long and Short Papers)}}, \bibinfo{pages}{4171--4186} (\bibinfo{publisher}{Association for Computational Linguistics}, \bibinfo{address}{Minneapolis, Minnesota}, \bibinfo{year}{2019}).
    \newblock \urlprefix\url{https://aclanthology.org/N19-1423/}.

    \bibitem{chang2023muse}
    \bibinfo{author}{Chang, H.} \emph{et~al.}
    \newblock \bibinfo{title}{Muse: Text-to-image generation via masked generative transformers}.
    \newblock \emph{\bibinfo{journal}{arXiv preprint arXiv:2301.00704}}  (\bibinfo{year}{2023}).

    \bibitem{bach2015pixel}
    \bibinfo{author}{Bach, S.} \emph{et~al.}
    \newblock \bibinfo{title}{On pixel-wise explanations for non-linear classifier decisions by layer-wise relevance propagation}.
    \newblock \emph{\bibinfo{journal}{PloS one}} \textbf{\bibinfo{volume}{10}}, \bibinfo{pages}{e0130140} (\bibinfo{year}{2015}).

    \bibitem{horst2019explaining}
    \bibinfo{author}{Horst, F.}, \bibinfo{author}{Lapuschkin, S.}, \bibinfo{author}{Samek, W.}, \bibinfo{author}{M{\"u}ller, K.-R.} \& \bibinfo{author}{Sch{\"o}llhorn, W.~I.}
    \newblock \bibinfo{title}{Explaining the unique nature of individual gait patterns with deep learning}.
    \newblock \emph{\bibinfo{journal}{Scientific reports}} \textbf{\bibinfo{volume}{9}}, \bibinfo{pages}{2391} (\bibinfo{year}{2019}).

    \bibitem{anders2021software}
    \bibinfo{author}{Anders, C.~J.}, \bibinfo{author}{Neumann, D.}, \bibinfo{author}{Samek, W.}, \bibinfo{author}{Müller, K.-R.} \& \bibinfo{author}{Lapuschkin, S.}
    \newblock \bibinfo{title}{Software for dataset-wide xai: From local explanations to global insights with {Zennit}, {CoRelAy}, and {ViRelAy}}.
    \newblock \emph{\bibinfo{journal}{arXiv preprint arXiv:2106.13200}}  (\bibinfo{year}{2021}).

    \bibitem{kim2024enhancing}
    \bibinfo{author}{Kim, M.} \& \bibinfo{author}{Park, S.}
    \newblock \bibinfo{title}{Enhancing accuracy and convenience of golf swing tracking with a wrist-worn single inertial sensor}.
    \newblock \emph{\bibinfo{journal}{Scientific Reports}} \textbf{\bibinfo{volume}{14}}, \bibinfo{pages}{9201} (\bibinfo{year}{2024}).

    \bibitem{murray2020international}
    \bibinfo{author}{Murray, A.} \emph{et~al.}
    \newblock \bibinfo{title}{International consensus statement: methods for recording and reporting of epidemiological data on injuries and illnesses in golf}.
    \newblock \emph{\bibinfo{journal}{British journal of sports medicine}} \textbf{\bibinfo{volume}{54}}, \bibinfo{pages}{1136--1141} (\bibinfo{year}{2020}).

    \bibitem{robinson2019systematic}
    \bibinfo{author}{Robinson, P.~G.} \emph{et~al.}
    \newblock \bibinfo{title}{Systematic review of musculoskeletal injuries in professional golfers}.
    \newblock \emph{\bibinfo{journal}{British journal of sports medicine}} \textbf{\bibinfo{volume}{53}}, \bibinfo{pages}{13--18} (\bibinfo{year}{2019}).

    \bibitem{robinson2024prospective}
    \bibinfo{author}{Robinson, P.~G.} \emph{et~al.}
    \newblock \bibinfo{title}{A prospective study of injuries and illnesses among 910 amateur golfers during one season}.
    \newblock \emph{\bibinfo{journal}{BMJ Open Sport \& Exercise Medicine}} \textbf{\bibinfo{volume}{10}}, \bibinfo{pages}{e001844} (\bibinfo{year}{2024}).

    \bibitem{glazier2024ecological}
    \bibinfo{author}{Glazier, P.~S.}
    \newblock \bibinfo{title}{An ecological-dynamical approach to golf science: implications for swing biomechanics, club design and customisation, and coaching practice}.
    \newblock \emph{\bibinfo{journal}{Sports Biomechanics}} \textbf{\bibinfo{volume}{23}}, \bibinfo{pages}{2467--2488} (\bibinfo{year}{2024}).

    \bibitem{egret2003analysis}
    \bibinfo{author}{Egret, C.}, \bibinfo{author}{Vincent, O.}, \bibinfo{author}{Weber, J.}, \bibinfo{author}{Dujardin, F.} \& \bibinfo{author}{Chollet, D.}
    \newblock \bibinfo{title}{Analysis of 3d kinematics concerning three different clubs in golf swing}.
    \newblock \emph{\bibinfo{journal}{International journal of sports medicine}} \textbf{\bibinfo{volume}{24}}, \bibinfo{pages}{465--470} (\bibinfo{year}{2003}).

    \bibitem{joyce2013three}
    \bibinfo{author}{Joyce, C.}, \bibinfo{author}{Burnett, A.}, \bibinfo{author}{Cochrane, J.} \& \bibinfo{author}{Ball, K.}
    \newblock \bibinfo{title}{Three-dimensional trunk kinematics in golf: between-club differences and relationships to clubhead speed}.
    \newblock \emph{\bibinfo{journal}{Sports Biomechanics}} \textbf{\bibinfo{volume}{12}}, \bibinfo{pages}{108--120} (\bibinfo{year}{2013}).

    \bibitem{Van_Wouwe_2024_CVPR}
    \bibinfo{author}{Van~Wouwe, T.}, \bibinfo{author}{Lee, S.}, \bibinfo{author}{Falisse, A.}, \bibinfo{author}{Delp, S.} \& \bibinfo{author}{Liu, C.~K.}
    \newblock \bibinfo{title}{Diffusionposer: Real-time human motion reconstruction from arbitrary sparse sensors using autoregressive diffusion}.
    \newblock In \emph{\bibinfo{booktitle}{Proceedings of the IEEE/CVF Conference on Computer Vision and Pattern Recognition (CVPR)}}, \bibinfo{pages}{2513--2523} (\bibinfo{year}{2024}).

\end{thebibliography}
\end{document}